\documentclass[11pt]{article}
\usepackage{coling2018}
\usepackage{times}
\usepackage{url}
\usepackage{latexsym}

\usepackage{booktabs}
\usepackage[normalem]{ulem}
\usepackage{booktabs}
\usepackage{multirow}
\usepackage{color}
\usepackage{soul}

\usepackage{graphicx}
\usepackage{caption}
\usepackage{subcaption}
\graphicspath { {figures/} }
\usepackage{verbatim}

\newcommand{\COMMENT}[1] {}

\newcommand{\fix}[2]{\st{}\textcolor{black}{#2}}

\title{A Comparison of Transformer and Recurrent Neural Networks \\ on Multilingual Neural Machine Translation}

\author{Surafel M. Lakew \\
  University of Trento\\Fondazione Bruno Kessler \\
  {\tt lakew@fbk.eu} \\\And
  Mauro Cettolo \\
  Fondazione Bruno Kessler \\
  {\tt cettolo@fbk.eu} \\\And
  Marcello Federico \\
  MMT Srl, Trento\\Fondazione Bruno Kessler \\
  {\tt federico@fbk.eu} \\}

\date{}

\begin{document}
\maketitle
\begin{abstract} 
Recently, neural machine translation (NMT) has been extended to multilinguality, that is to handle more than one translation direction with a single system. Multilingual NMT showed competitive performance against pure bilingual systems. Notably, in low-resource settings, it proved to work effectively and efficiently, thanks to 
shared representation space that is forced across languages and induces a sort of transfer-learning.
Furthermore, multilingual NMT enables so-called zero-shot inference across language pairs never seen at training time.  
Despite the increasing interest in this framework, an in-depth analysis of what a multilingual NMT model is capable of and what it is not is still missing.   
Motivated by this, our work (i)~provides a quantitative and comparative analysis of the translations produced by bilingual, multilingual and zero-shot systems; 
(ii)~investigates the translation quality of two of the currently dominant neural architectures in MT, which are the Recurrent and the Transformer ones; and (iii)~quantitatively explores how the closeness between languages influences the zero-shot translation.
Our analysis leverages multiple professional post-edits of automatic translations by several different systems and focuses both on automatic standard metrics (BLEU and TER) and on widely used error categories, which are lexical, morphology, and word order errors. 
\end{abstract}

\section{Introduction}\label{sec:intro}
%
%
\blfootnote{
    %
    %
    %
    %
    %
    %
    \hspace{-0.65cm}  
    This work is licensed under a Creative Commons Attribution 4.0 International License. License details: \url{http://creativecommons.org/licenses/by/4.0/}
}
As witnessed by recent machine translation evaluation campaigns (IWSLT~\cite{iwslt-overview:2017}, WMT~\cite{bojar-EtAl:2017:WMT1}), in the past few years several model variants and training procedures have been proposed and tested in neural machine translation (NMT). NMT models were mostly employed in conventional single language-pair settings, where the training process exploits a parallel corpus from a source language to a target language, and the inference involves only those two languages in the same direction. However, there have also been attempts to incorporate multiple languages in the source~\cite{luong2015multi,zoph2016multi,lee2016fully}, in the target~\cite{dong2015multi}, or in both sides like \newcite{firat2016multi} which combines a shared attention mechanism and multiple encoder-decoder layers. Regardless, the simple approach proposed in \newcite{johnson2016google} and \newcite{ha2016toward} remains outstandingly effective: it relies on  single ``universal'' encoder, decoder and attention modules, and manages multilinguality by introducing an artificial token at the beginning of the input sentence to specify the requested target language.

The current NMT state-of-the-art includes the use of recurrent neural networks, initially introduced in~\newcite{sutskever2014sequence,cho2014learningGRU}, convolutional neural networks, proposed by~\newcite{gehring2017convolutional}, and so-called transformer neural networks, recently proposed by~\newcite{vaswani2017attention}. All of them implement an encoder-decoder architecture, suitable for sequence-to-sequence tasks like machine translation,
and an attention mechanism~\cite{bahdanau2014neural}.

Besides specific studies focusing on new architectures and modules, like \newcite{luong2015effective} that empirically evaluates different implementations of the attention mechanism, the comprehension of what a model can learn and the errors it makes has been drawing much attention of the research community, as evidenced by the number of recent publications aiming at comparing the behavior of neural vs. phrase-based systems~\cite{bentivogli2016neural,toral2017multifaceted,bentivogli2018neural}.
However, understanding the capability of multilingual NMT models in general and zero-shot translation, in particular, has not been thoroughly analyzed yet. By taking the bilingual model as the reference, this work quantitatively analyzes the translation outputs of multilingual and zero-shot models, aiming at answering the following research questions: 

\begin{itemize}
\setlength{\parskip}{0.5mm}
\item How do bilingual, multilingual, and zero-shot systems compare in terms of general translation quality? Is there any translation aspect better modeled by each specific system?
\item How do Recurrent and Transformer architectures compare in terms of general translation quality? Is there any translation aspect better modeled by each specific system?
\item What is the impact of using related languages data in training a zero-shot translation system for a given language pair?
\end{itemize}

To address these questions, we exploit the data collected in the IWSLT $2017$ MT evaluation campaign~\cite{iwslt-overview:2017} and made publicly available by the organizers. The campaign was the first featuring a multilingual shared MT task, spanning five languages (English, Dutch, German, Italian, and Romanian) and all their twenty possible translation directions. 
In addition to the official external single reference of the test sets, we can also rely on professional post-edits of the outputs of nine Romanian$\rightarrow$Italian and of nine Dutch$\rightarrow$German participants' systems. Hence, we exploit the availability of multiple Italian and German references to perform a thorough analysis for identifying, comparing and understanding the errors made by different neural system/architectures we are interested in; in particular, we consider pairs of both related  languages (Romanian$\rightarrow$Italian, Dutch$\rightarrow$German) and unrelated languages (Romanian$\rightarrow$German and Dutch$\rightarrow$Italian). Furthermore, to explore the impact of using data from other related languages, French and Spanish are considered for training purposes as well, in particular for analyzing the behavior of zero-shot ${x}$$\rightarrow$Italian systems, ${x}$ representing any source language distant from Italian.

In the following sections, we begin with a brief review of related work on quantitative analysis of MT tasks ($\S$\ref{sec:related_work}). Then, we give an overview of NMT ($\S$\ref{sec:nmt}) with a contrast between the Recurrent ($\S$\ref{subsec:nmt_rnn}) and Transformer ($\S$\ref{subsec:nmt_tranformer}) approaches, and a summary on multilingual and zero-shot translation ($\S$\ref{subsec:m-nmt}). Section ~($\S$\ref{sec:exp}), describes the dataset and preprocessing pipeline ($\S$\ref{subsec:dataset}), qualitative evaluation data ($\S$\ref{subsec:eval_data}), experimental setting ($\S$\ref{subsec:settings}), models ($\S$\ref{subsec:models}) and the evaluation methods ($\S$\ref{subsec:eval}). In Section ($\S$\ref{sec:analysis}), we analyze the overall translation quality for related and unrelated language directions. Before the summary and conclusion, we will focus on lexical, morphological and word-order error types for the fine-grained analysis ($\S$\ref{subsec:fined-grained}).

\section{Related Work}\label{sec:related_work}
Recent trends in NMT evaluation show that post-editing helps to identify and address the weakness of systems~\cite{bentivogli2018neural}. Furthermore, the use of multiple post-edits in addition to the manual reference is gaining more and more ground~\cite{bentivogli2016neural,koehn2017six,toral2017multifaceted,bentivogli2018neural}.  For our investigation, we follow the error analysis approach defined in \newcite{bentivogli2018neural}, where multiple post-edits are exploited in order to quantify morphological, lexical, and word order errors, a simplified error classification with respect to that proposed in~\newcite{vilar2006error}, which settles two additional classes, namely missing and extra words.

The first work that compares bilingual, multilingual, and zero-shot systems comes from the IWSLT $2017$ evaluation campaign~\cite{iwslt-overview:2017}. The authors analyze the outputs of several systems through two human evaluation methods: direct assessment which focuses on the generic assessment of overall translation quality, and post-editing which directly measures the utility of a given MT output to translators. Post-edits are also exploited to run a fine-grained analysis of errors made by the systems. The main findings are that (i)~a single multilingual system is an effective alternative to a bunch of bilingual systems, and that (ii)~zero-shot translation is a viable solution even in low-resource settings. Motivated by those outcomes, in this work we explore in more detail the practical feasibility of multilingual and zero-shot approaches. In particular, we explore the benefit of adding training data involving related languages in a zero-shot setting and, in that framework, we compare the behavior of state-of-the-art Transformer and Recurrent NMT models.

\section{Neural Machine Translation}\label{sec:nmt}
A standard state-of-the-art NMT system comprises of an encoder, a decoder and an attention mechanism, which are all trained with maximum likelihood in an end-to-end fashion~\cite{bahdanau2014neural}. Although there are different variants of the encoder-attention-decoder based approach, this work focuses on the Recurrent LSTM-based variant~\cite{sutskever2014sequence} and the Transformer model~\cite{vaswani2017attention}.

In both the Recurrent and Transformer approaches, the encoder is purposed to cipher a source sentence into hidden state vectors, whereas the decoder uses the last representation of the encoder to predict symbols in the target language. In a broad sense, the attention mechanism improves the prediction process by deciding which portion of the source sentence to emphasize at a time \cite{luong2015effective}. In the following two subsections, we briefly summarize the two architecture types.

\subsection{Recurrent NMT}\label{subsec:nmt_rnn}
In this case, the source words are first mapped to vectors with which the encoder recurrent network is fed. When the $<$eos$>$ (i.e. end of sentence) symbol is seen, the final time step initializes the decoder recurrent network. At each time step of the decoding, the attention mechanism is applied over the encoder hidden states and combined with the current hidden state of the decoder to predict the next target word. Then, the prediction is fed back to the decoder (i.e. input feeding), to predict the next word, until the $<$eos$>$ symbol is generated~\cite{sutskever2014sequence,cho2014learningGRU}.

\subsection{Transformer NMT}\label{subsec:nmt_tranformer}
The Transformer architecture works by relying on a self-attention (\textit{intra-attention}) mechanism, removing all the recurrent operations that are found in the previous approach.
In other words, the attention mechanism is repurposed to compute the latent space representation of both the encoder and the decoder sides.
However, with the absence of recurrence, \textit{positional-encoding} is added to the input and output embeddings. Similarly, as the time-step in a recurrent network, the positional information provides the Transformer network with the order of input and output sequences. In our work, we use the absolute positional encoding, but very recently the use of the relative positional information has been shown to improve performance~\cite{shaw2018self}. The model is organized as a stack of encoder-decoder networks that works in an auto-regressive way, using the previously generated symbol as input for the next prediction. Both the decoder and encoder can be composed of uniform layers, each built of two sub-layers, i.e., a multi-head self-attention layer and a position wise feed-forward network (FFN) layer.
The multi-head sub-layer enables the use of multiple attention functions with a similar cost of utilizing attention, while the FFN sub-layer is a fully connected network used to process the attention sublayers; as such, FFN applies two linear transformations on each position and a ReLU~\cite{vaswani2017attention}.

\subsection{Multilingual NMT}\label{subsec:m-nmt}
Recent efforts in multilingual NMT using a single encoder-decoder and an attention mechanism~\cite{johnson2016google,ha2016toward} have shown to improve translation performance with minimal complexity.
Multilingual NMT models can be trained with parallel corpora of several language pairs in \textit{many-to-one, one-to-many}, or \textit{many-to-many} translation directions. The main idea that distinguishes multilingual NMT training and inference from a single language pair NMT is that in preprocessing, a \emph{language-flag} is appended to the source side of each segment pair. The flag specifies the target language the source is paired with at training time. Moreover, it enables a zero-shot inference by directing the translation to a target language never seen at training time paired with the source. In addition to reducing training and maintenance complexity of several single language pair systems, the two main advantages of multilingual NMT is the performance gain for low-resource languages, and the possibility to perform a zero-shot translation. 

However, the translations generated by multilingual and zero-shot systems have not been investigated in detail yet. This includes analyzing how the model behaves solely relying on a ``language-flag'' as a way to redirect the inference. Recent works have shown that the target language-flag is weaker in a low-resource language setting~\cite{lakew2017multilingual}. Thus, in addition to analyzing the behavior of bilingual and multilingual models, mainly, the zero-shot task requires a careful investigation.

\section{Data and Experiments}\label{sec:exp}
\subsection{Datasets and preprocessing}\label{subsec:dataset}
The experimental setting comprises seven languages; for each language pair, we use the $\approx$200,000 parallel sentences made publicly available by the IWSLT 2017 evaluation campaign~\cite{iwslt-overview:2017}, partitioned in training, development, and test sets. 
In the preprocessing pipeline, the raw data is first tokenized and cleaned by removing empty lines. Then, a shared byte pair encoding (BPE) model~\cite{sennrich2015sub-word} is trained using the union of the source and target sides of the training data. The number of BPE segmentation rules is set to $8,000$, following the suggestion of \newcite{denkowski2017stronger} for experiments in small training data condition. 
For the case of Transformer training, the internal sub-word segmentation~\cite{wu2016google} provided by the Tensor2Tensor library\footnote{https://github.com/tensorflow/tensor2tensor} is used. Note that prepending the ``language-flag'' on the source side of the corpus is specific to the multilingual and zero-shot models.

\begin{table}[!t]
\centerline{
\begin{tabular}{l|rrrrrr}
\multicolumn{1}{c}{}  & encoder-decoder & embedding & hidden & encoder & decoder & batch \\
\multicolumn{1}{c}{}  & type    & size      & units  & depth   & depth   & size \\
  \hline
Recurrent & LSTM & 512 & 1024 & 4 & 4 & 128 seg \\
Transformer & Self-Attention & 512 & 512 & 6 & 6 & 2048 tok \\
\bottomrule
\end{tabular}}
\caption{\label{table:parameters} Hyper-parameters used to train Recurrent and Transformer models, unless differently specified.}
\end{table}

\subsection{Evaluation data}\label{subsec:eval_data}
For our investigation, we exploit the nine post-edits available from the IWSLT 2017 evaluation campaign. Post-editing regarded the bilingual, multilingual, and zero-shot runs of three different participants to the two  
tasks Dutch (Nl)$\rightarrow$German (De) and Romanian (Ro)$\rightarrow$Italian (It).
Human evaluation was performed on a subset ($603$ sentences) of the nine runs, involving professional translators.
Details on data preparation and the post-editing task can be found in~\newcite{iwslt-overview:2017}. 

The translation directions we consider in this work are Nl/Ro$\rightarrow$De and Nl/Ro$\rightarrow$It. The choice of \textit{German} and \textit{Italian} as the target languages is motivated by (i)~the availability of multiple post-edits for the fine-grained analysis and (ii)~the possibility of varying the linguistic distance between the source and the target languages, allowing experimental configurations suitable to answer the research questions raised in Section~$\S$\ref{sec:intro}.   

As said, for Nl$\rightarrow$De and Ro$\rightarrow$It the human evaluation sets consist of $603$ segments. Since post-editing involved only those two language pairs, for the other two directions considered in this work, namely Nl$\rightarrow$It and Ro$\rightarrow$De, 
we tried to exploit at best the available post-edits by looking for all and only the segment pairs of the Nl$\rightarrow$It and Ro$\rightarrow$De tst$2017$ sets for which the target side exactly matches (at least) one of the segment pairs of the Ro$\rightarrow$It and Nl$\rightarrow$De human evaluation sets. This way, we were able to find $478$ matches on the Italian sides and $444$ on the German sides, which therefore become the sizes of the human evaluation sets of Ro$\rightarrow$De and Nl$\rightarrow$It, respectively, for which we can re-use the available post-edits. 

It is worth to note that in general, the post-edits from the evaluation campaign are not actual post-edits of MT outputs generated in our experiments, with some exceptions discussed later, therefore they should rather be considered as multiple external references.

\subsection{Training setting}\label{subsec:settings}
Each of the three system types, namely bilingual, multilingual and zero-shot, is trained using both Recurrent and Transformer architectures, with the proper training data provided in the IWSLT 2017 evaluation campaign. Meta training parameters were set in a preliminary stage with the aim of maximizing the quality of each approach. Recurrent NMT experiments are carried out using the open source OpenNMT-py\footnote{https://github.com/OpenNMT/OpenNMT-py}~\cite{klein2017opennmt}, whereas the Transformer models are trained using the Tensor2Tensor toolkit. Hence, we took the precaution of selecting the optimal training and inference parameters for both approaches and toolkits. For instance, for our low-resource setting characterized by a high data sparsity, the dropout~\cite{srivastava2014dropout} is set to $0.3$~\cite{gal2016theoreticallyDropout} in Recurrent models and to $0.2$ in Transformer models to prevent over-fitting. Similarly, Adam~\cite{kingma2014adam} optimizer with an initial learning rate of either $0.001$ (RNN) or $0.2$ (Transformer) is used. If the perplexity does not decrease on the validation set or the number of epochs is above $7$, a learning rate decay of $0.7$ is applied in the Recurrent case. For the Transformer case, the learning rate is increased linearly in the early stages (\emph{warmup\_training\_steps=$16000$}); after that, it is decreased with an inverse square root of training step~\cite{vaswani2017attention}. Table~\ref{table:parameters} summarizes the list of hyper-parameters.

\subsection{Models}\label{subsec:models}
To address the research questions listed in Section~\ref{sec:intro}, we train five types of models using either the Recurrent or the Transformer approaches. All models are trained up to convergence, eventually the best performing checkpoint on the dev set is selected. Table~\ref{tab:models} summarizes the systems tested in our experiments. As references, we consider four bilingual systems (in short NMT) trained on the following directions:  
Nl$\rightarrow$De/It and Ro$\rightarrow$De/It. The first term of comparison 
is a many-to-many multilingual system (in short M-NMT) trained in all directions in the set \{En,De,Nl,It,Ro\}. Then, we  test zero-shot translation (ZST) between related languages, namely Nl$\rightarrow$De and Ro$\rightarrow$It, by training a multilingual NMT without any data for these language pairs. We also test zero-shot translation between unrelated languages (ZST\_A), namely Ro$\rightarrow$De and Nl$\rightarrow$It, by excluding parallel data between these languages. Finally, for the same unrelated zero-shot directions we  also train multi-lingual systems (ZST\_B) that include data related to Romanian and Italian, namely En$\leftrightarrow$Fr/Es.

\begin{table}[!t]
\small
\centering
\begin{tabular}{p{1.4cm}p{1.6cm}p{11.9cm}}
Model	&	 \#Directions	 &	 System description	 \\ \midrule
NMT	&	1	&	\textit{Four pure bilingual models for the Nl$\rightarrow$De/It and Ro$\rightarrow$De/It directions}	\\
M-NMT          &    20      &	\textit{Multilingual, trained on all directions in the set \{En,De,Nl,It,Ro\}}	\\
ZST     &  16      &	\textit{Zero-shot, trained as multilingual but removing also Nl$\leftrightarrow$De and It$\leftrightarrow$Ro data}    \\
\midrule 
ZST\_A & 12      & \textit{Zero-shot, trained as ZST but removing also De$\leftrightarrow$Ro and Nl$\leftrightarrow$It data}	\\
ZST\_B & 16      &	\textit{Zero-shot, trained as ZST\_A but adding En$\leftrightarrow$Fr/Es data} \\
\bottomrule
\end{tabular}
\caption{\label{tab:models} The training setting of $4$*bilingual, $1$*multilingual, and $3$*zero-shot systems.}
\end{table}

\subsection{Evaluation methods}\label{subsec:eval}
Systems are compared in terms of  BLEU~\cite{papineni2002bleu} (as implemented in \emph{multi-bleu.perl}~\footnote{A script from the Moses SMT toolkithttp://www.statmt.org/moses}) and TER~\cite{Snover:06} scores, on the single references of the official IWSLT test sets.

\noindent
In addition, two TER-based scores are reported, namely the multiple-reference TER (mTER) and a lemma-based TER (lmmTER), which are instead computed on the nine post-edits of the IWSLT 2017 human evaluation set. In mTER, TER is computed by counting, for each segment of the MT output, the minimum number of edits across all the references and dividing by the average length of references. lmmTER is computed similarly to mTER but looking for matches at the lemma level instead of surface forms. Significance tests for all scores are reported using Multeval~\cite{clark2011better} tool. 

\noindent
Systems are also compared in terms of three well known and widely used error categories, that is lexical, morphological, and word order errors, exploiting TER and post-edits as follows. First, the MT outputs and the corresponding post-edits are lemmatized and POS-tagged; for that, we used ParZu~\cite{Sennrich2013parZu} for German and TreeTagger~\cite{Schmid1994treeTagger} for Italian. Then, the lemmatized outputs are evaluated against the corresponding post-edits via a variant of the {\it tercom} implementation\footnote{Available at \url{wit3.fbk.eu/show.php?release=2016-02&page=subjeval}} of TER: in addition to computing TER, the tool provides complete information about matching lemmas, as well as shift (matches after displacements), insertion, deletion, and substitution operations. Since for each lemma the tool keeps track of the corresponding original word form and POS tag, we are able to measure the number of errors falling in the three error categories, following the scheme described in detail in~\newcite{bentivogli2018neural}.

\section{Translation Analysis}\label{sec:analysis}
\subsection{Related languages}
\begin{table}[!t]
\centering
\small
\begin{tabular}{llrrrrrrrr}
\toprule
\multicolumn{1}{l}{\multirow{2}{*}{Direction}} & \multicolumn{1}{l}{\multirow{2}{*}{System}} & \multicolumn{4}{c}{Recurrent}                                                               & \multicolumn{4}{c}{Transformer}                    \\ \cmidrule(l){3-6}  \cmidrule(l){7-10} 
\multicolumn{1}{c}{}                           & \multicolumn{1}{r}{}                        & BLEU  & TER                        & \multicolumn{1}{r}{mTER} & \multicolumn{1}{r|}{lmmTER} & BLEU  & TER                        & \multicolumn{1}{r}{mTER} & \multicolumn{1}{r}{lmmTER} \\ \midrule
\multirow{3}{*}{Nl$\rightarrow$De}             & \multicolumn{1}{l|}{NMT}                    & 18.05                    & \multicolumn{1}{r|}{ 64.61} & 23.70 & \multicolumn{1}{r|}{ 20.60} & \bf 18.37     & \multicolumn{1}{r|}{\bf 63.74} & 27.95 & 23.86  \\
                                               & \multicolumn{1}{l|}{M-NMT}                  & 17.79                    & \multicolumn{1}{r|}{66.18} & 21.75 & \multicolumn{1}{r|}{18.28} & \bf $^{\uparrow}$19.95 & \multicolumn{1}{r|}{\bf 61.90} & 23.62 & 20.05  \\
                                               & \multicolumn{1}{l|}{ZST}                    & 17.06                    & \multicolumn{1}{r|}{65.73} & 26.35 & \multicolumn{1}{r|}{22.29} & \bf $^{\uparrow}$19.13 & \multicolumn{1}{r|}{\bf62.69} & \bf25.19 & \bf21.53  \\ \midrule
\multirow{3}{*}{Ro$\rightarrow$It}             & \multicolumn{1}{l|}{NMT}                    & 22.16                    & \multicolumn{1}{r|}{59.35} & 22.99 & \multicolumn{1}{r|}{20.39} & \bf 22.48     & \multicolumn{1}{r|}{\bf57.34} & 26.60 & 23.36  \\
                                               & \multicolumn{1}{l|}{M-NMT}                  & 21.69                    & \multicolumn{1}{r|}{59.50} & 21.12 & \multicolumn{1}{r|}{18.46} & \bf $^{\uparrow}$22.12 & \multicolumn{1}{l|}{\bf57.51} & 25.05 & 21.57  \\
                                               & \multicolumn{1}{l|}{ZST}                    & 18.72                    & \multicolumn{1}{r|}{62.08} & 29.66 & \multicolumn{1}{r|}{26.15} & \bf $^{\uparrow}$21.29 & \multicolumn{1}{r|}{\bf59.08} & \bf26.93 & \bf23.33  \\ \bottomrule
\end{tabular}
\caption{\label{table:bleu_1} Automatic scores on tasks involving related languages. BLEU and TER are computed on  test$2017$, while mTER and lmmTER are reported for human evaluation sets. Best scores of the Transformer model against the Recurrent are highlighted in bold, whereas arrow $^{\uparrow}$ indicates statistically significant differences ($p<0.05$).}
\end{table}

\noindent
First, we compare the bilingual (NMT), multilingual (M-NMT), and zero-shot (ZST) systems on the two tasks \textit{Nl$\rightarrow$De} and \textit{Ro$\rightarrow$It}, implemented as either Recurrent or Transformer networks, in terms of automatic metrics. As stated above, BLEU and TER exploit the official external reference of the whole test sets, while  mTER and lmmTER are utilize the multiple post-edits of the (smaller) IWSLT human evaluation test set. Scores are given in Table~\ref{table:bleu_1}.

\noindent
Looking at the BLEU/TER scores, it is evident that Transformer performs  better in all the three model variants. In particular, for the multilingual and the zero-shot models, the gain is statistically significant. On the contrary, the mTER and lmmTER scores are better for the Recurrent architecture; in this case, the outcome is misleading since the nine post-edits include those generated by correcting the outputs of the three Recurrent systems. As such, the translations of the Recurrent systems are rewarded over the  translations produced by the Transformer systems, thus making the comparison not fair.

\noindent
As far as the models are compared, the bilingual one is the best in three out of four cases, the exception being the Transformer/Nl$\rightarrow$De. Nonetheless, it is worth to note the good performance of the multilingual model in terms of mTER and lmmTER. This result holds true in both Recurrent and Transformer approaches, regardless of the BLEU score. 
We hypothesize that the main reason behind this is the higher number of linguistic phenomena observed in training, thanks to the use of data from multiple languages, which makes the multilingual models more \emph{robust} than the bilingual models.

\subsection{Unrelated languages}
\begin{table}[!t]
\small
\centering
\begin{tabular}{llrrrrrrrr}
\toprule
\multicolumn{1}{l}{\multirow{2}{*}{Direction}} & \multicolumn{1}{l}{\multirow{2}{*}{System}} & \multicolumn{4}{c}{Recurrent}                                                                     & \multicolumn{4}{c}{Transformer}                                                            \\ \cmidrule(l){3-6}  \cmidrule(l){7-10}
\multicolumn{1}{r}{}                           & \multicolumn{1}{r}{}                        & BLEU  & TER                        & \multicolumn{1}{r}{mTER} & \multicolumn{1}{r|}{lmmTER} & BLEU  & TER                        & \multicolumn{1}{r}{mTER} & \multicolumn{1}{r}{lmmTER} \\ \midrule
\multirow{3}{*}{Ro$\rightarrow$De}             & \multicolumn{1}{l|}{NMT}                    & 13.99 & \multicolumn{1}{r|}{72.70} & 61.82                    & \multicolumn{1}{r|}{54.61}  & \bf $^{\uparrow}$16.52 & \multicolumn{1}{r|}{\bf66.71} & \bf55.68                    & \bf48.44                      \\
                                               & \multicolumn{1}{r|}{ZST\_A}                 & 14.93 & \multicolumn{1}{r|}{69.38} & 58.26                    & \multicolumn{1}{r|}{51.08}  & \bf $^{\uparrow}$16.46 & \multicolumn{1}{r|}{\bf66.88} & \bf54.72                    & \bf48.25                      \\
                                               & \multicolumn{1}{l|}{ZST\_B}                 & 14.75 & \multicolumn{1}{r|}{69.29} & 58.26                    & \multicolumn{1}{r|}{51.37}  & \bf $^{\uparrow}$16.55 & \multicolumn{1}{r|}{\bf67.18} & \bf55.29                    & \bf48.03                      \\ \midrule
\multirow{3}{*}{Nl$\rightarrow$It}             & \multicolumn{1}{l|}{NMT}                    & 18.88 & \multicolumn{1}{r|}{63.79} & 58.79                    & \multicolumn{1}{r|}{52.16}  & \bf $^{\uparrow}$20.22 & \multicolumn{1}{r|}{\bf60.88} & \bf55.52                    & \bf48.56                      \\
                                               & \multicolumn{1}{l|}{ZST\_A}                 & 18.77 & \multicolumn{1}{r|}{62.97} & 58.80                    & \multicolumn{1}{r|}{51.32}  & \bf $^{\uparrow}$19.80 & \multicolumn{1}{r|}{\bf60.24} & \bf54.06                    & \bf47.16                      \\
                                               & \multicolumn{1}{l|}{ZST\_B}                 & 18.87 & \multicolumn{1}{r|}{62.40} & 57.34                    & \multicolumn{1}{r|}{50.17}  & \bf $^{\uparrow}$20.61 & \multicolumn{1}{r|}{\bf59.41} & \bf53.04                    & \bf46.17                      \\ \bottomrule
\end{tabular}
\caption{\label{table:bleu_ru_2} Evaluation results for the unrelated language directions. BLEU and TER scores are computed with single references, while mTER and lmmTER are computed with nine post-edits. Best scores of the Transformer over the corresponding Recurrent architectures are highlighted in bold, whereas arrow $^{\uparrow}$ indicates statistically significant differences ($p<0.05$).}
\end{table}

In unrelated language directions, our experimental setting is aimed at evaluating the impact of source \emph{language-relatedness} with the target. Particularly, we focus on the zero-shot setup given its intrinsic difficulty, by taking the bilingual systems as references. 
Table~\ref{table:bleu_ru_2} provides BLEU and TER based scores for the Ro$\rightarrow$De and Nl$\rightarrow$It directions. 

\noindent
Concerning the ZST\_A training condition, in one case (Recurrent Ro$\rightarrow$De) it outstandingly allows to outperform the pure bilingual system, while in the other cases there is no significant difference between ZST\_A and NMT, proving once again that zero-shot translation built on the ``language-flag'' of M-NMT is really effective~\cite{johnson2016google}: in fact,  at most a slight performance degradation is recorded as the number of pairs used in training decreases~\cite{lakew2017multilingual}.
Although gains are rather limited, adding training data involving Romance target languages (French and Spanish, ZST\_B) close to Italian impacts as hoped: ZST\_B scores are in general better than both NMT and ZST\_A in Nl$\rightarrow$It, while they do not degrade with respect to ZST\_A in Ro$\rightarrow$De.

\noindent
Similarly to what is observed for related pairs (Table~\ref{table:bleu_1}), the Transformer architecture shows definitely higher quality than the RNN one, confirming the capability of the approach to infer unseen directions.
The overall outcomes from Tables~\ref{table:bleu_1} and \ref{table:bleu_ru_2} are: (i)~multilingual systems have the potential to effectively model the translation either in zero-shot or non zero-shot conditions; (ii)~zero-shot translation is a viable option to enable translation without training samples;
(iii)~the Transformer is the best performing approach, particularly in the zero-shot directions. 

\noindent
The next section is devoted to a fine-grained analysis of errors made by the various systems at hand, with the aim of assessing the outcomes based on automatic metrics.

\section{Fine-grained Analysis}\label{subsec:fined-grained}
Following the error classification defined in Section~\ref{subsec:eval}, now we focus on lexical, morphological, and reordering error distributions to characterize the behavior of the three types of models and the two sequence-to-sequence learning approaches considered in this work.

\noindent
As anticipated in the previous section, it is expected that scores computed with reference to post-edits penalize Transformer over Recurrent systems because the outputs of the latter were post-edited, but not those of the former. We try to mitigate this bias by relying on the availability of multiple post-edits which likely allows to better match the Transformer runs than having a single reference would do. For the fine-grained analysis, we use instead the expedient of computing error distributions that are normalized with respect to the error counts observed in a bilingual reference system. In the next two sections, the fine-grained analysis is reported for related and unrelated languages pairs, consecutively.

\subsection{Related languages}
\begin{table}[!t]
\small
\centering
\begin{tabular}{lrrrrr|rrrrr}
\toprule 
\multicolumn{1}{l}{\multirow{2}{*}{Nl$\rightarrow$De}} & \multicolumn{5}{c}{\bf Recurrent}                         & \multicolumn{5}{c}{\bf Transformer}                 \\ \cmidrule(l){2-6}  \cmidrule(l){7-11} 
\multicolumn{1}{c}{}                       & NMT   & M-NMT & $\Delta_{NMT}$ & ZST   & $\Delta_{NMT}$ & NMT   & M-NMT & $\Delta_{NMT}$ & ZST   & $\Delta_{NMT}$ \\ \midrule
Lexical                                    & 77.29 & 69.65 & -7.64     & 83.73 & +6.44       & 76.47 & 64.83 & -11.64    & 69.53 & -6.94      \\
Morph                                      & 15.41 & 16.51 & +1.10       & 19.1 & +3.69       & 15.70 & 13.96 & -1.74     & 14.13 & -1.57     \\
Reordering                                 & 5.53  & 3.14  & -2.39     & 5.41  & -0.12      & 6.20  & 4.97  & -1.23     & 5.41  & -0.79    \\
Morph. \& Reo.                               & 1.76  & 1.02  & -0.74     & 1.61  & -0.15      & 1.63  & 1.36  & -0.27     & 1.53  & -0.10      \\ \midrule
Total                                      & 100   & 90.31   & -9.69     & 109.84   & +9.84       & 100   & 85.12   & \bf -14.88  & 90.6   & \bf -9.40     \\  \bottomrule
\end{tabular}
\caption{Distribution of lexical, morphological, and reordering error types from the two MT approaches. Reported values are normalized with respect to the total error count of the respective bilingual reference model (NMT). $\Delta_{NMT}$ are variations with respect to the bilingual reference models (NMT).}
\label{table:fine-grained_rnn_t_nlde}
\end{table}

Table~\ref{table:fine-grained_rnn_t_nlde} provides the distribution over the error types by the bilingual (NMT), multi-lingual (M-NMT), and zero-shot (ZST) models, implemented either with Recurrent or Transformer architectures, for the Nl$\rightarrow$De translation direction. We also report, for each error type and M-NMT and ZST system, the observed relative difference of errors with respect to the bilingual reference model (NMT).

\noindent
Considering each error category, we observe the same general trend for all systems: the lexical errors represent by far the most frequent category (76-77\%), followed by morphology (15-\fix{18}{16}\%) and reordering (3-6\%) errors; cases of words whose morphology and positioning are both wrong, represent about 1-2\% of the total errors. Beyond the similar error distributions, it is worth to note the variation of errors made by M-NMT and ZST models with respect to those of the NMT model: for the Recurrent architecture, there is a decrease of $9.69$ and an increase of $9.84$ points, respectively. On the contrary, the Transformer architecture yields improvements for both models: total errors reduce by $14.88$ and $9.40$ points, respectively. The result for the Transformer ZST system is particularly valuable since the average error reduction comes from remarkable improvements across all error categories.

\noindent
For the Ro$\rightarrow$It direction, results are given in Table~\ref{table:fine-grained_rnn_t_roit}. 
Although to a different extent, we observe a picture similar to that of Nl$\rightarrow$De discussed above: lexical errors is the type of error committed to a greater extent, multilingual models outperform their bilingual correspondents (more for the Recurrent than for the Transformer models), and ZST is competitive with bilingual NMT only if the Transformer architecture is adopted.

\begin{table}[!t]
\small
\centering
\begin{tabular}{lrrrrr|rrrrr}
\toprule
\multicolumn{1}{l}{\multirow{2}{*}{Ro$\rightarrow$It}} & \multicolumn{5}{c}{Recurrent}                             & \multicolumn{5}{c}{Transformer}                    \\ \cmidrule(l){2-6} \cmidrule(l){7-11} 
\multicolumn{1}{c}{}                       & NMT   & M-NMT & $\Delta_{NMT}$     & ZST   & $\Delta_{NMT}$ & NMT   & M-NMT & $\Delta_{NMT}$ & ZST   & $\Delta_{NMT}$    \\ \midrule
Lexical                                    & 80.63 & 73.81 & -6.82          & 102.79 & +22.16      & 81.97 & 76.01 & -5.96      & 84.12 & +2.15          \\
Morph                                      & 12.33 & 12.86 & +0.53           & 16.00 & +3.67      & 11.49 & 11.79 & +0.30       & 12.44 & +0.95          \\
Reordering                                 & 5.74  & 3.71 & -2.03         & 6.09  & +0.35       & 5.35  & 4.64  & -0.71     & 4.81 & -0.54        \\
Morph. \& Reo.                               & 1.30  & 1.15  & -0.15        & 2.18  & +0.88      & 1.19  & 1.09  & -0.10      & 1.09  & -0.10         \\ \midrule
Total                                      & 100   &  91.54  & \bf -8.46 & 127.07  & +27.07      & 100   & 93.52   & -6.48      & 102.45   & \bf+2.45 \\ \bottomrule
\end{tabular}
\caption{Distribution of the error types in the Ro$\rightarrow$It direction for the Recurrent and Transformer approaches. From the variation of errors that compare M-NMT and ZST models with the bilingual reference (NMT), a larger margin of error is observed in case of Transformer ZST model.}
\label{table:fine-grained_rnn_t_roit}
\end{table}

\begin{table}[!t]
\small
\centering
\begin{tabular}{lrrrrr|rrrrr}
\toprule
\multicolumn{1}{l}{\multirow{2}{*}{Ro$\rightarrow$It}} & \multicolumn{5}{c}{Recurrent}                   &  \multicolumn{5}{c}{Transformer}           \\ \cmidrule(lr){2-6} \cmidrule(l){7-11} 
\multicolumn{1}{c}{}                   & NMT    & ZST\_A & $\Delta_{NMT}$ & ZST\_B & $\Delta_{NMT}$ & NMT & ZST\_A & $\Delta_{NMT}$ & ZST\_B & $\Delta_{NMT}$ \\ \midrule
Lexical                            &   80.63     & 108.27  & +27.64      & 100.31  & +19.68   &  81.97  &   82.11  & +0.14       & 76.76  & -5.21      \\
Morph                                &  12.33    & 17.11  & +4.78     & 17.23  & +4.90      & 11.49   & 13.09  & +1.60      & 11.59  & +0.10       \\
Reordering                           &  5.74    & 6.20  & +0.46       & 6.16   & +0.42       & 5.35  &   5.18    & -0.17      & 5.59   & +0.24      \\
Morph. \& Reo.                       &   1.30     & 2.22   & +0.92      & 2.30   & +1.00     & 1.19  &   1.16   & -0.03     & 1.02   & -0.17     \\ \cmidrule(r){1-11} 
Total                               &   100    & 133.81    & +33.81      & 126    & +26.00   & 100   &  101.53   & \bf+1.53       & 94.96    & \bf-5.04      \\ \bottomrule
\end{tabular}
\caption{Error distribution of\fix{bilingual (NMT),}{} ZST\_A and ZST\_B models for the Recurrent and Transformer variants. Transformer achieves the highest error reduction, particularly in the ZST\_B model setting.}
\label{table:related_zst_b_ro-it}
\end{table}

\noindent
Training under the zero-shot conditions ZST\_A and ZST\_B assume less training data available and permit to measure the impact of introducing additional parallel data from related languages. We considered training conditions ZST\_A and ZST\_B here to perform Ro$\rightarrow$It  zero shot translation and report the outcomes in Table~\ref{table:related_zst_b_ro-it}.

\noindent
Results show error counts for each condition normalized with respect to the 
corresponding bilingual reference models (NMT). The most interesting aspect comes from the fact that  global variations in the normalized error counts of the zero-shot translation can be here associated with the relatedness and variety of languages in the training data. As recently reported~\cite{lakew2017multilingual}, zero-shot performance of Recurrent models in a low resource setting seems highly associated with the number of languages provided in the training data. This is also confirmed by comparing performance of Recurrent models across the ZST (Table~\ref{table:fine-grained_rnn_t_roit}), ZST\_A and ZST\_B conditions. In particular, variations from the bilingual reference model, show significant degradation when some language directions are removed (from +27.07 to +33.81) and a significant improvement when two related languages are added (from +33.81 to +26.00). Remarkably, the Transformer zero-shot model seems less sensitive to the removal or addition of languages: actually a slight improvement is observed after removing  Nl$\rightarrow$It and De$\rightarrow$Ro (ZST\_A), i.e., from +2.45 to +1.53, followed by a large improvement when En$\rightarrow$Fr/Es (ZST\_B)
are added, i.e. from +1.53 to -5.04. Notice that the latter results outperform the bilingual model. Overall, across all experiments, we see slight changes in the distribution of errors types. On the other hand, increases or drops of specific error types with respect to the bilingual reference model show sharper differences across the different conditions.
For instance, the best performing Transformer model (ZST\_B in Table~\ref{table:related_zst_b_ro-it}) seems to gain over the reference bilingual systems only in terms of lexical errors (-5.21). The zero-shot Transformer model trained under the ZST condition (Table~\ref{table:fine-grained_rnn_t_roit}) although globally worse than the bilingual reference, seems instead slightly better than the reference concerning reordering error (-0.54), which account for 5.35\% of the total number of errors.

\subsection{Unrelated languages}
In our second scenario, we evaluate the relative changes in the error distribution for the unrelated  language directions (Ro$\rightarrow$De and Nl$\rightarrow$It). This section complements the translation results reported in Table~\ref{table:bleu_ru_2}, analyzing the runs from the ZST\_A and ZST\_B models in a different manner.  

\begin{table}[!t]
\small
\centering
\begin{tabular}{lrrrrr|rrrrr}
\toprule
\multicolumn{1}{l}{\multirow{2}{*}{Ro$\rightarrow$De}} & \multicolumn{5}{c}{Recurrent}                           & \multicolumn{5}{c}{Transformer}                   \\ \cmidrule(l){2-6}  \cmidrule(l){7-11}
\multicolumn{1}{c}{}                       & NMT   & ZST\_A & $\Delta_{NMT}$ & ZST\_B & $\Delta_{NMT}$ & NMT   & ZST\_A & $\Delta_{NMT}$ & ZST\_B & $\Delta_{NMT}$ \\ \midrule
Lexical                                    & 79.18 & 74.42  & -4.76      & 74.09  & -5.09      & 79.21 & 79.11  & -0.10      & 78.52  & -0.69      \\
Morph                                      & 9.91  & 10.35  & +0.44      & 10.07  & 0.16       & 9.92  & 10.05  & +0.13       & 10.87  & +0.95       \\
Reordering                                 & 7.33  & 6.16  & -1.17       & 6.16   & -1.17     & 7.19  & 6.88   & -0.31      &  7.22  & +0.03       \\
Morph. \& Reo.                             & 3.58  & 3.47   & -0.11      & 3.47   & -0.11      & 3.68  & 3.52   & -0.16      & 3.60   & -0.08      \\ \midrule
Total                                      & 100  & 94.4    & \bf-5.60   & 93.79    & \bf-6.21      & 100  & 99.55    & -0.45      & 100.21    & +0.21       \\ \bottomrule
\end{tabular}
\caption{Error distribution of the bilingual (NMT), ZST\_A and ZST\_B model runs for the unrelated Ro$\rightarrow$De direction. The Transformer moder shows the smallest sensitivity to the change in the number of training language pairs.}
\label{table:unrelated_zst_a-b_ro-de}
\end{table}

In the Ro$\rightarrow$De unrelated direction (Table~\ref{table:unrelated_zst_a-b_ro-de}), the Recurrent model shows a reduction in the error rate of $5.60$ points (ZST\_A) and $6.21$ points (ZST\_B) with respect to the bilingual (NMT) reference model, while for the Transformer no significant differences are observed. These results confirm what observed in the automatic evaluation on the reference translations (Table~\ref{table:bleu_ru_2}). The gain observed by the Recurrent model on the ZST\_B condition is mainly on lexical (-4.76 points) and reordering errors (-1.17 points) is probably due to the poor performance 
of its bilingual counterpart.

\begin{table}[!t]
\centering
\small
\begin{tabular}{lrrrrr|rrrrr}
\toprule
\multicolumn{1}{l}{\multirow{2}{*}{Nl$\rightarrow$It}} & \multicolumn{5}{c}{Recurrent}                                                                 & \multicolumn{5}{c}{Transformer}                           \\ \cmidrule(l){2-6} \cmidrule(l){7-11} 
\multicolumn{1}{c}{}                       & \multicolumn{1}{l}{NMT} & \multicolumn{1}{l}{ZST\_A} & $\Delta_{NMT}$ & ZST\_B & $\Delta_{NMT}$ & NMT   & ZST\_A & $\Delta_{NMT}$     & ZST\_B & $\Delta_{NMT}$    \\ \midrule
Lexical			& 81.08       & 80.7     & -0.38     & 78.79  	& -2.29    	& 81.15 	& 79.36  	& -1.79         & 77.48  & -3.67        \\
Morph       	& 8.47        & 9.03     & +0.56     & 8.38  	& -0.09		& 9.01  	& 9.2   	& +0.19         & 9.03   & +0.02        \\
Reordering  	& 7.78        & 6.63     & -1.15     & 6.32   	& -1.46     & 7.51  	& 6.74 		& -0.77         & 6.51   & -1.00        \\
Morph \& Reo 	& 2.67        & 2.54	 & -0.13     & 2.38   	& -0.29    	& 2.34  	& 2.41   	& +0.07         & 2.45   & +0.11        \\ \midrule
Total           & 100         & 98.89    & -1.11     & 95.86  	& -4.14  	& 100   	& 97.71    	& \bf-2.29		& 95.47    & \bf-4.53 	\\ \bottomrule
\end{tabular}
\caption{Error distribution of the bilingual (NMT), ZST\_A and ZST\_B model runs. $\Delta_{NMT}$ shows the relative change in the error distribution of the zero-shot models with respect to the bilingual reference models.}
\label{table:unrelated_zst_a-b_nl-it}
\end{table}

As far as the the Nl$\rightarrow$It unrelated direction (Table~\ref{table:unrelated_zst_a-b_nl-it}) is concerned, both Recurrent and Transformer ZST models show to reduce the error counts over the bilingual reference model. Actually, a similar trend occurs in Ro$\rightarrow$De (Table~\ref{table:unrelated_zst_a-b_ro-de}), but with a relatively higher error reduction in case of the Transformer model. In particular, the Transformer model shows reductions of $-2.29$ points for ZST\_A and $-4.53\%$ for ZST\_B, whereas for the Recurrent model the improvements are slightly lower, namely $-1.11$ (ZST\_A) and $-4.14$ (ZST\_B) points. Remarkably,  both the Recurrent and Transformer models benefit from additional training data related to Italian (compare ZST\_A and ZST\_B). 

In conclusion, we observe that error counts of the zero-shot models in unrelated directions can increase (Table~\ref{table:unrelated_zst_a-b_ro-de}) when compared to the bilingual model. However, in the related language direction the most interesting aspect is observed with the discount of error in the Nl$\rightarrow$It direction (Table~\ref{table:unrelated_zst_a-b_nl-it}). In particular, the ZST\_B zero-shot model showed $>$$2.0\%$ error reduction over the ZST\_A model. This gain is directly related to the newly introduced training data (i.e., English$\leftrightarrow$French/Spanish) in case of ZST\_B.

\section*{Summary and Conclusions}
In this work, we showed how bilingual, multilingual, and zero-shot models perform in terms of overall translation quality, as well as the errors types produced by each system. Our analysis compared Recurrent models with the recently introduced Transformer architecture. Furthermore, we explored the impact of grouping related languages for a zero-shot translation task. In order to make  the overall evaluation more sound, BLEU and TER scores were complemented with mTER and lmmTER, leveraging multiple professional post-edits. Our investigation on the translation quality and the results of the fine-grained analysis shows that:

\begin{itemize}
\item Multilingual models consistently outperform bilingual models with respect to all considered  error types, i.e., lexical, morphological, and reordering.
\item The Transformer approach delivers the best performing multilingual models, with a larger gain over corresponding bilingual models than observed with RNNs.
\item Multilingual models between related languages achieve the best performance scores and relative gains over corresponding bilingual models.
\item When comparing zero-shot and bilingual models, relatedness of the source and target languages does not play a crucial role.
\item The Transformer model delivers the best quality in all considered zero-shot condition and translation directions.
\end{itemize}

Our fine-grained analysis looking at three types of errors (lexical, reordering, morphology) show significant differences in the error distributions 
across the different translation directions, even when switching the source language with another source language of the same family. No particular differences in the error distributions were observed across neural MT architectures (Recurrent vs. Transformer), while some marked differences were observed when comparing bilingual, multilingual, and zero-shot systems. A more in-depth analysis of these differences will be carried out in future work.

\section*{Acknowledgements}
This work has been partially supported by the EC-funded projects ModernMT (H2020 grant agreement no. 645487) and QT21 (H2020 grant agreement no. 645452). We also gratefully acknowledge the support of NVIDIA Corporation with the donation of the Titan Xp GPU used for this research.

\bibliographystyle{acl}
\bibliography{coling2018}

\end{document}